\documentclass[10pt,twocolumn,letterpaper]{article}

\usepackage{iccv}
\usepackage{times}
\usepackage{epsfig}
\usepackage{graphicx}
\usepackage{amsmath}
\usepackage[normalem]{ulem}
\usepackage{amssymb}
\usepackage{graphicx}
\usepackage{amsmath}
\usepackage{multirow}
\usepackage{amssymb}
\usepackage{colortbl}
\usepackage{multicol}
\usepackage{booktabs}
\usepackage{tabularx}
\usepackage[export]{adjustbox}

\newcommand{\mn}{LLNeRF}

\newcommand{\cc}{\mathbf{c}}
\newcommand{\rr}{\mathbf{r}}

\newcommand{\pixel}[1]{{#1}_r}
\newcommand{\pixeli}[1]{{#1}_{ri}}
\newcommand{\pixelj}[1]{{#1}_{rj}}
\newcommand{\point}[1]{{#1}}

\newcommand{\ray}{r}
\newcommand{\RR}{\mathbf{R}}
\newcommand{\dd}{\mathbf{d}}

\newcommand{\illu}{\mathbf{v}}
\newcommand{\Illu}{\mathbf{V}}
\newcommand{\xx}{\mathbf{x}}
\newcommand{\GG}{\gamma}
\newcommand{\EE}{\mathbb{E}}
\newcommand{\cch}{\mathbf{\hat c}}
\newcommand{\ccb}{\mathbf{\bar c}}
\newcommand{\cgt}{\mathbf{\tilde c}}
\newcommand{\ryn}[1]{\textcolor[rgb]{0,0,1}{#1}}
\newcommand{\hlb}[1]{\textcolor[rgb]{0,0.5,0.5}{#1}}

\newcommand{\cropH}{0.1\textwidth}
\newcommand{\addimg}[1]{
\adjustimage{height=\cropH}{images/#1}
}

\newcommand{\EQref}[1]{Eq.~\ref{#1}}

\renewcommand{\ryn}[1]{\textcolor[rgb]{0,0,0}{#1}}
\renewcommand{\hlb}[1]{\textcolor[rgb]{0,0,0}{#1}}

\newcommand{\hll}[1]{\textcolor[rgb]{0,0,0}{#1}}

\usepackage[svgnames]{xcolor}
\definecolor{lightblue}{RGB}{46,150,222}
\usepackage[pagebackref=true,breaklinks=true,letterpaper=true,citecolor=lightblue,colorlinks,bookmarks=false]{hyperref}

\usepackage[capitalize]{cleveref}
\usepackage{multirow}

\crefname{section}{Sec.}{Secs.}
\Crefname{section}{Section}{Sections}
\Crefname{table}{Table}{Tables}
\crefname{table}{Tab.}{Tabs.}

\iccvfinalcopy 

\def\iccvPaperID{5161} 

\ificcvfinal\fi

\begin{document}


\title{Lighting up NeRF via Unsupervised Decomposition and Enhancement}

\author{Haoyuan Wang\textsuperscript{1}, Xiaogang Xu\textsuperscript{2}, Ke Xu\textsuperscript{1}, Rynson W.H. Lau\textsuperscript{1}\\
\textsuperscript{1}Department of Computer Science, City University of Hong Kong \\ 
\textsuperscript{2} Zhejiang Lab, Zhejiang University\\
{\small \url{https://onpix.github.io/llnerf}}
}

\maketitle
\ificcvfinal\thispagestyle{empty}\fi






\begin{abstract}
Neural Radiance Field (NeRF) is a promising approach for synthesizing novel views, given a set of images and the corresponding camera poses of a scene. However, images photographed from a low-light scene can hardly be used to train a NeRF model to produce high-quality results, due to their low pixel intensities, heavy noise, and color distortion. Combining existing low-light image enhancement methods with NeRF methods also does not work well due to the view inconsistency caused by the individual 2D enhancement process. In this paper, we propose a novel approach, called Low-Light NeRF (or LLNeRF), to enhance the scene representation and synthesize normal-light novel views directly from sRGB low-light images \hll{in an unsupervised manner}. The core of our approach is a decomposition of radiance field learning, which allows us to enhance the illumination, reduce noise and correct the distorted colors jointly with the NeRF optimization process. Our method is able to produce novel view images with proper \hll{lighting} and vivid colors and details, given a collection of camera-finished low dynamic range (8-bits/channel) images from a low-light scene. Experiments demonstrate that our method outperforms existing low-light enhancement methods and NeRF methods. 
\vspace{-4mm}
\end{abstract}

\section{Introduction}
\label{sec:intro}

Neural Radiance Field (NeRF)~\cite{nerf} is a powerful approach to render novel view images through learning scene representations as implicit functions. 
These implicit functions are parameterized by multi-layer perceptrons (MLPs) and optimized by measuring the colorimetric errors of the input views.
Consequently, high-quality input images are the precondition for the high-quality results of NeRF.
In other words, training NeRF models typically requires the input images to have high visibility, and almost all the pixels \ryn{to} faithfully represent the scene illumination and object colors.
However, when taking photos under low-light conditions, the quality of the images is not guaranteed.
Low-light images typically have low visibility. Noise from the camera is also relatively amplified due to the low photons, which further buries the scene details and distorts object colors. Such characteristics of low-light images fail existing NeRF models in producing high-quality novel view images.

\begin{figure}[t]
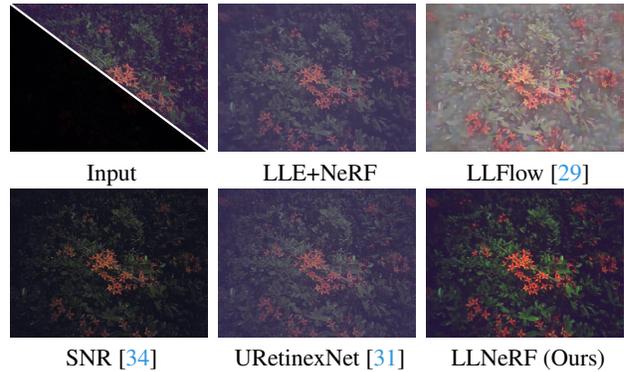

\small
\renewcommand{\tabcolsep}{1pt}
\renewcommand{\cropH}{0.15\textwidth}
\renewcommand{\addimg}[1]{
\adjustimage{width=\cropH}{images/teaser/#1}
}
\centering
\begin{tabular}{ccc}

  \addimg{cmp3_1__input.png} 
& \addimg{cmp3_1__baseline.png}
& \addimg{cmp3_1__llflow_all.png}
\\
Input & LLE+NeRF & LLFlow~\cite{llflow} 
\\
  \addimg{cmp3_1__sdsd_outdoor.png} 
& \addimg{cmp3_1__uretinex_all.png}
& \addimg{cmp3_1__ours_f3.png}
\\
SNR~\cite{Xu_2022_SNRA} & URetinexNet~\cite{Wu_2022_URetinex} & {\mn} (Ours)
\end{tabular}
\vspace{-2mm}
\caption{
A comparison of the baseline model (LLE+NeRF), SOTA low light enhancement models, and our model. 
}
\label{fig:teaser}
\vspace{-4mm}
\end{figure}

We note that recently there are some methods proposed to train NeRF models from degraded inputs~\cite{rawnerf,hdrnerf,ma2022deblur}.
Ma~\etal~\cite{ma2022deblur} present a method to synthesize novel view images from blurry inputs taken in normal-light scenes.
Mildenhall~\etal~\cite{rawnerf} show that when training with high dynamic range RAW data, NeRF can be robust to zero-mean noise of low-light input images.
Huang~\etal~\cite{hdrnerf} propose HDR-NeRF, which produces high dynamic range (HDR) novel views from a set of low dynamic range (LDR) input images taken at different known exposure levels.
The latter two methods take advantages of HDR information and metadata (\ie, exposure levels) recorded in the RAW images to enhance the scene representations.
However, these methods do not work \ryn{on} camera-finished sRGB images (8-bits/channel) taken in low-light scenes. Unlike RAW data, sRGB images are produced by the camera ISP process. They are of low dynamic range and low signal-to-noise ratio. 

A straightforward solution to this problem is to first enhance the low-light input images and then use the enhanced results to train a NeRF model.
However, while this may be able to improve the brightness, existing low-light enhancement models do not consider how to maintain consistency across multi-view images. Besides, these learning-based enhancement methods tend to learn specific mappings of brightness from their own training data, which may not generalize well to in-the-wild scenes.
%
These two reasons cause NeRF to learn biased information across different views due to the \ryn{view-dependent} optimization of NeRF, \ryn{resulting in} unrealistic novel images. See examples in \cref{fig:teaser}.

In this paper, we propose a new approach for rendering novel normal-light images from a set of 8-bit low-light sRGB images without the supervision of ground truth.
Our key \hll{solution} to this problem
\hll{is that: the colors of 3D points can be decoupled into view-dependent and view-independent components within the NeRF optimization, and the view-dependent component is dominated by the effect of lighting. So the manipulations of the lighting-related view-independent components are able to enhance the brightness, correct the colors, and reduce the noise while keeping the texture and structure of the scene.}
Experiments demonstrate that the proposed method outperforms the state-of-the-art NeRF models and the baselines (\ie, combining NeRF with state-of-the-art enhancement methods). 

In summary, we propose the first method to reconstruct a NeRF model of proper lighting from a collection of LDR low-light images. Our main contributions \ryn{includes}:
\begin{enumerate}
    \item \hll{We propose to decompose NeRF into view-dependent and -independent color components for enhancement. The decomposition does not require ground truth. }
    \item We formulate an unsupervised method to enhance the lighting and correct the colors while rendering noise-free novel view images.
    \item We collect a real-world dataset, and conduct extensive experiments to analyze our method and demonstrate its effectiveness in real-world scenes. 
\end{enumerate}

\section{Related Work}

{\flushleft\bf Neural Radiance Field} represents 3D scenes via parameterized implicit functions and allows to render high-quality novel view images.
However, NeRF is sensitive to the input images as it relies on the colorimetric optimization of the input images.
\ryn{Some} methods focus on improving the robustness of NeRF to dynamic scenes in the wild by using, \eg, time-of-flight data~\cite{attal2021torf}, latent appearance modelling~\cite{martin2021nerf}, camera self-calibration~\cite{jeong2021self}, depth estimation~\cite{wei2021nerfingmvs,deng2022depth}, and semantic labels~\cite{zhi2021place}.

\ryn{Some other} methods~\cite{rawnerf,hdrnerf,ma2022deblur} propose to train NeRF models from degraded inputs.
Ma~\etal~\cite{ma2022deblur} propose a deformable sparse kernel module for deblurring while synthesizing novel view images from blurry inputs.
Mildenhall~\etal~\cite{rawnerf} propose to train NeRF directly on camera raw images for handling the low visibility and noise of low-light scenes.
Huang~\etal~\cite{hdrnerf} proposes the HDR-NeRF to synthesize novel view HDR images from a collection of LDR images of different exposure levels, which implicitly handles the exposure fusion using a tone mapper.
Unlike the above methods, in this paper, we aim to address the problem of training NeRF using a group of low-light sRGB images, which is more challenging due to \ryn{the low visibility, low dynamic ranges, large noise, and high color distortions}.

{\flushleft\bf Low-light Enhancement} aims to improve the content visibility of images taken from low-light scenes.
A line of deep enhancement methods learns specific mappings from low-light images to \ryn{expert-retouched images or images} captured with high-end cameras. These methods \ryn{propose}
different priors and techniques \ryn{aiming} to enhance the capacity of neural networks for learning such mappings, \eg, using HDR information~\cite{hdrnet,Yang-cvpr18-drht,sharma2021nighttime}, generative adversarial learning~\cite{Ignatov-iccv17-dslr,deep-photo-enhancer,enlighten-gan,ren-tip19-lowlight}, deep parametric filters~\cite{deelpf}, and reinforcement learning~\cite{Park-cvpr18-distort-and-recover,yu-nips18-deepexposure}.
Some methods propose to decompose the images into illumination and detail layers~\cite{Cai-tip18-SICE}, layers of different frequency components~\cite{xu2020learning}, and regions of different exposures~\cite{msec,wang2022lcdp} for enhancement.
Recently, Xu~\etal~\cite{Xu_2022_SNRA} propose to combine transformer and CNNs to model long-range correlations for low-light enhancement.

Our work is closer in spirit to the Retinex-based enhancement methods~\cite{Zhang-mm18-HQexposure,Cai-iccv17-JieP,deepupe,zhang-acmmm19-kindling,ruas, Wu_2022_URetinex}. These methods first decompose the input image into \hll{the illumination and reflectance} layers and then enhance the illumination layer of the image.
While these methods learn such decomposition from 2D images, which typically lack geometry information, our method works in the radiance field, resulting in a more realistic decomposition and enhancement.

\section{Preliminary Knowledge and Analysis}

We first \ryn{summarize} how neural radiance field (NeRF) works under normal-light scenes and then explain the challenges \ryn{for NeRF to handle} low-light scenes.


\subsection{NeRF Preliminary}

Given a set of posed training images, NeRF~\cite{nerf} learns to render the color of every single pixel $\pixel{\cc}$ for a ray $r$, which could be uniquely identified by the camera index 
and the 2D pixel coordinates.
NeRF represents a scene by a radiance field, which takes as input an arbitrary single ray cast $\ray(t) = \mathbf{o} + t\dd$, where $\mathbf{o}, \dd, t$ are the ray origin, ray direction, and the distance along the ray, respectively. 
The rendering process has three steps: (1) NeRF samples $n$ points along the ray $\ray(t)$, \ie, $t_i \in \mathbf{t}$ where $\mathbf{t}$ is a $n$-D vector, between the near and far image planes using the hierarchical sampling strategy;
(2) NeRF applies an optional transform function $\psi(\cdot)$ to the sampled \ryn{coordinate} vector $\mathbf{t}$ along the ray;
%
and (3) NeRF uses the MLPs $F_{\text{density}}, F_{\text{color}}$ to learn the volume density and the color along the rays, denoted by $\point{\sigma}$ and $\point{\cc}$, from 
$\mathbf{t}$
and the view direction $\dd$ as:
\begin{equation}
\begin{cases}
    (\tau, \point{\sigma}) =  F_{\text{density}}(\psi({\ray(t_i)}), \dd; \Theta_{F_{\text{density}}}), \; \;t_i \in \mathbf{t} \\
    \point{\cc} =  F_{\text{color}}(\tau, \dd; \Theta_{F_{\text{color}}}),
\end{cases}
\label{eq:mlp}
\end{equation}
where $\tau$ is the intermediate features learned by the neural network. Different NeRF implementations may have different versions of the transform function $\psi(\cdot)$. The original NeRF implementation~\cite{nerf} uses the frequency positional encoding function as $\psi(\cdot)$, while in Mip-NeRF~\cite{mipnerf}, $\psi(\cdot)$ is implemented as interval splitting and integrated positional encoding. In this paper, we use \hlb{the implementation of } Mip-NeRF~\cite{mipnerf}, and the \ryn{pixel colors} are rendered as:
\begin{equation}
    \pixel{\cc} = \sum_i w_i\point{\cc}_i = \sum_i \left(1-e^{-\point{\sigma}\delta_i}\right) e^{-\sum_{j<i} \sigma_j\delta_j} \point{\cc}_i,
    \label{eq:vr}
\end{equation}
where $\delta_i = t_{i+1}-t_i$. $\pixel{\cc}$ is the final rendered 3-channel pixel color of the corresponding ray $\ray(t)$. NeRF is then optimized under the supervision of the ground-truth pixel \ryn{colors} $\pixel{\cgt}$ of the training images.

\subsection{Challenges}

Since the NeRF model directly optimizes its implicit radiance field according to the 2D projected images, training a NeRF model using low-light sRGB images has two challenges.
First, NeRF cannot handle the low pixel intensity of low-light images, and can only produce dark images as novel views.
Second, although~\cite{rawnerf} shows that NeRF is robust to zero-mean noise in the raw domain due to its essential integration process, the signal-to-noise ratio \ryn{of the camera-finished sRGB images is much lower than that of} the raw images. In addition, the camera ISP process changes the linearity property of raw images and blends scene radiance with noise together in the camera-finished sRGB images. Hence, NeRF is not able to handle noise and color distortion when training on low-light sRGB images.

To obtain a normal-light NeRF, combining low-light enhancement methods with NeRF (LLE+NeRF) may be a possible solution.
However, as existing low-light enhancement methods mainly learn a mapping from low light to normal light based on specific training data. This mapping may not generalize well to new scenes that are out of the distributions of the training data.
Hence, using images enhanced by these existing methods to train a NeRF model may produce low-quality novel view images.
On the other hand, taking multi-view images of both low-light and normal light at the same time as training data is not practical. 

In this work, we aim to develop a method to \hll{produce high-quality novel view images from low-light scenes in an unsupervised manner.}

\section{Our Unsupervised Approach}

\begin{figure*}
    \centering
    \makebox[\linewidth][c]{\includegraphics[width=1\linewidth]{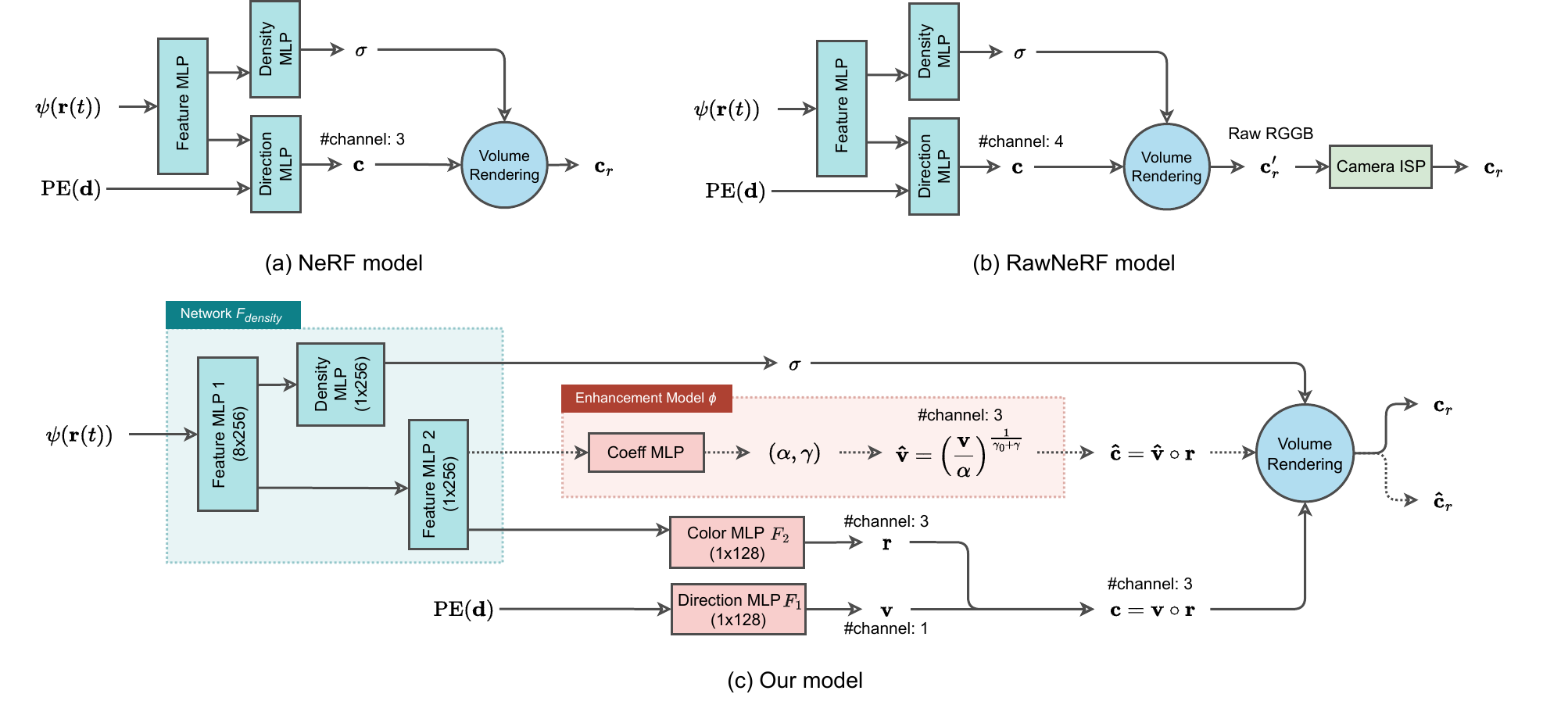}}
    \vspace{-7mm}
    \caption{
    The illustration of the \ryn{NeRF~\cite{nerf} model (a), RawNeRF~\cite{rawnerf} model (b), and our proposed model (c)}. The data flow of \ryn{our} unsupervised enhancement is shown \ryn{inside the} dashed line. Our model jointly learns the novel view images and enhances the \ryn{output}
    of all samples along the ray. Each final enhanced pixel is rendered using the volume rendering equation \ryn{as shown} in \cref{eq:vr}. 
    }
    \label{net}
    \vspace{-3mm}
\end{figure*}

The \ryn{main idea} of our work is to decompose the implicit radiance field of NeRF and then leverage priors to enhance the lighting, reduce noise and correct the colors of the \ryn{novel-view} images. \ryn{\cref{net}(c) shows the pipeline of our method.}

\begin{figure}[t]
    \centering
    \makebox[\linewidth][c]{\includegraphics[width=0.9\linewidth]{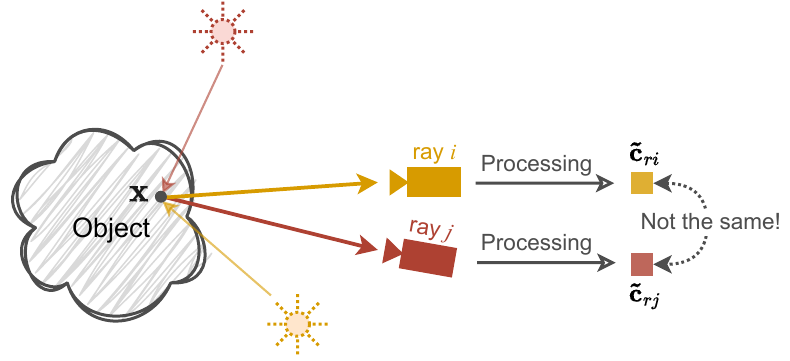}}
    \vspace{-0mm}
    \caption{
    The 2D projection $\pixeli{\cgt}, \pixelj{\cgt}$ of the same spatial point $\xx$ is not exactly identical but in the same color spectrum. \hll{The variance of color across views, \ie, the view-dependent component of the observed color, is dominated by the effect of lighting.}
    }
        \label{fig:lr}
        \vspace{-3mm}
\end{figure}

\subsection{Neural Radiance Field Decomposition} \label{sec:lr}

\hll{As shown in \cref{fig:lr}, when one 3D point $\xx$ in a static scene is projected to two pixels ($\pixeli{\cgt}$ and $\pixelj{\cgt}$) of two views, the colors of two pixels may appear differently, as the object surface may not be isotropic and the lighting is not uniform.
However, the colors of these two pixels are still in the same range of the color spectrum.
This suggests that the color of one 3D point $\xx$ can be decomposed into a view-independent basis component and a view-dependent component. The view-independent basis component represents the intrinsic color, which determines the spectrum range of the color of $\xx$. \hll{The view-dependent component accounts for factors that may cause color differences across views (in most situations lighting is the dominant factor, which varies depending on the position and color of the light sources and the orientation of the surface at $\xx$).
}
}

\hll{
Inspired by this, we propose to decompose the color $\cc$ into the product of view-dependent component $\illu$ that captures the lighting-related component and its reciprocal component $\rr$ that represents the color basis.
We leverage NeRF to constrain $\illu$ to be view-dependent and further formulate it to be a single channel representation that focuses on the manipulation of lighting intensity.
}

Consider the rendering of \ryn{a} pixel $\pixel{\cc}$ of image $\mathbf{I}$ in \cref{eq:vr}. Since an arbitrary image pixel $\pixel{\cc}$ is the weighted accumulation of the view-dependent color of all $\{\point{\cc}_i\}_{i=1}^{n}$ along the ray, we decompose each $\point{\cc}$ along the ray into \hll{$\illu$ and $\rr$}, and learn to enhance the color as:
\begin{equation}
    \begin{cases}
    \point{\illu} =  F_1(\tau, \dd; \Theta_{F_1})
    \quad\text{and}\quad
    \point{\rr} =  F_2(\tau; \Theta_{F_2}),\\
    
    \point{\cc} = \point{\illu} \circ \point{\rr}
    \quad\text{and}\quad
    \point{\cch} = \phi(\point{\illu}) \circ \point{\rr},

    \end{cases}\label{eq:dec}
\end{equation}
where $\point{\cch}$ is the enhanced color, $\phi$ is an enhancement function parameterized by a neural network, and $\circ$ denotes the pixel-wise multiplication.
$F_1$ and $F_2$ are two MLPs. Thus, the enhanced image $\mathbf{I}_e$ can be obtained as:
\begin{equation}
    \mathbf{I}_e = \{\pixel{\cch}\}, \text{where }
    \pixel{\cch} = \sum_i
    w_i
    \phi(\point{\illu}_i) \circ \point{\rr}_i.
    \label{eq:point-lr}
\end{equation}

\hll{Such a method enables the model to learn a reasonable decomposition, which has a simple form but with strong constraints when the unenhanced colors $\cc$ are supervised across views. We further demonstrate the effectiveness of the decomposition design in \cref{sec:cmp}.}

{\flushleft \bf Differences to the Image-based Decomposition.} Image-based low-light enhancement methods~\cite{deepupe,retinex-net,Wu_2022_URetinex, zhang-acmmm19-kindling} typically \ryn{leverage} the Retinex theory to decompose an image $\mathbf{I}$ into the illumination map $\mathbf{L}$ and reflectance map $\RR$ as:
\begin{equation}
    \mathbf{I} = \mathbf{L} \circ \RR,
    \label{eq:lr-decomp}
\end{equation}
where $\RR$ is invariant to the lighting condition, affected by the material and intrinsic color of objects in an image, and $\mathbf{L}$ is the response of the \hll{illumination. Their decomposition is typically guided with the normal-light ground truth images during training. The enhanced image is obtained by:}
\begin{equation}
    \mathbf{I}_e = \phi(\mathbf{L}) \circ \RR,
    \label{eq:real-lr}
\end{equation}
where $\mathbf{I}_e$ is the enhanced image, and $\phi$ is the enhancement function \ryn{(\eg, the tone-mapping curve or a deep CNN), which is also supervised by GT}. 

\hll{In contrast, our method is unsupervised without ground truth for training. It works in the 3D neural radiance field with geometry information, and
leverages reasonable prior (\cref{fig:lr}) to constrain the decomposition process.
We compare the decomposition results of 2D-based method and ours in \cref{fig:lr-cmp}.}

\subsection{Unsupervised Enhancement}

\hll{In addition to the unsupervised decomposition, we propose an unsupervised enhancement method to enhance light up NeRF model. 
}

\subsubsection{Denoising}\label{denoising}

Let $\xx$ be a spatial point with a large density (\ie, the color of $\xx$ is dominant in the pixels) in a scene. \ryn{It} has multiple projections $C_\xx = \{\pixel{\cgt}\}$ in the training images.
We have $\pixel{\cgt} = \pixel{\ccb} + \mathbf{n}$, where $\pixel{\ccb}$ is the actual color and $\mathbf{n}$ is a small permutation noise sampled from an unknown distribution. During the training, the predicted color at $\xx$, \ie, $\cc_\xx$, is supervised by all pixels in $C_\xx$ and the gradients are propagated from different rays.

As the loss function of different rays is an unweighted average, the model tends to learn the smallest average deviation from the observations in $C_\xx$, and the learned $\pixel{\cc}$ would converge to the expectation of $\pixel{\cgt}$, \ie,
\begin{equation}
    \pixel{\cc} \approx \EE\{\pixel{\cgt}\} = \pixel{\ccb} + \EE\{\mathbf{n}\}.
    \label{eq:denoise}
\end{equation}
In RAW images, we could empirically assume the \ryn{noise in each training image is  zero-mean}~\cite{rawnerf}, \ie, $\EE\{\mathbf{n}\} = 0$. However, the non-linear processes applied to RAW images change the distribution of the raw \ryn{noise}, such that $\pixel{\cc}$ is converged to a biased value $\pixel{\ccb} + \EE\{\mathbf{n}\}$. Accordingly, the predicted colors along the ray $\point{\cc}$ would converge to $\point{\ccb} + \mathbf{b}$, where $\point{\ccb}$ is the ideal predicted color, and $\mathbf{b}$ is the bias introduced by the noise.

This indicates that the multi-view optimization of the implicit neural radiance field can still smooth the image and reduce the noise in our problem. 
\hll{However, applying this denoising scheme is not sufficient, as the converged pixel values may be biased, leading to color distortions. }
We introduce our color correction and enhancement method next.

\subsubsection{The Enhancement of $\illu$} \label{illu-enhance}

\begin{figure}[t]
    \centering
    \makebox[\linewidth][c]{\includegraphics[width=1.07\linewidth]{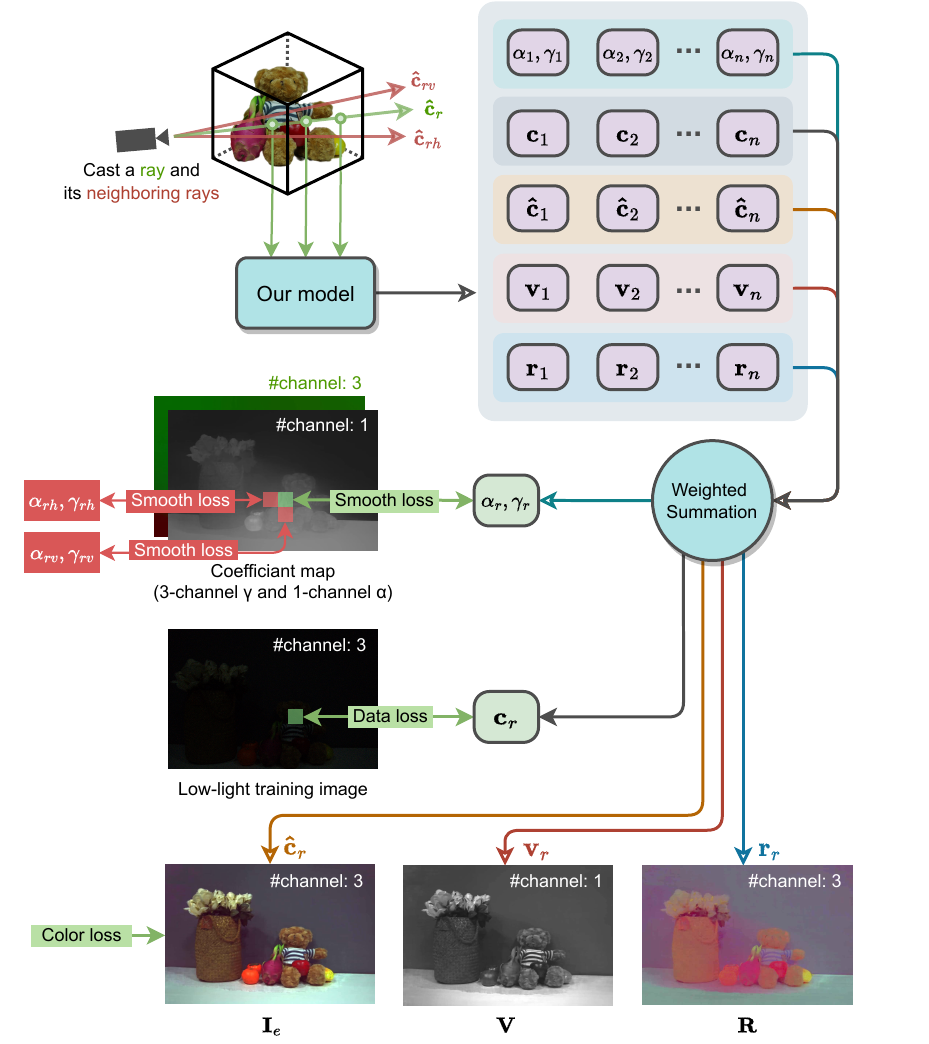}}
    \vspace{0em}
    \caption{
    Illustration of our training pipeline and the proposed loss functions. 
    The pixels are denoted as small blocks in green and red.
    }
    \label{fig:loss}
\end{figure}

We use \EQref{eq:point-lr} to enhance the $\point{\illu}$ along the ray for each spatial coordinate and view direction, \ie, $\mathbf{\hat v} = \phi(\point{\illu})$.
We propose to enhance $\point{\illu}$ using a dynamic gamma correction under the constraint of the rendered RGB value $\pixel{\cch}$, as: 
\begin{equation}
\mathbf{\hat v} = \phi(\point{\illu}) = \left( \point{\illu} \over \alpha \right)^{{1 \over \GG_0 + \GG}},
\label{eq:enhance}
\end{equation}
\hlb{
where $\alpha$ is a scalar and $\GG$ is a 3D vector. Both the two coefficients are the output of the enhancement network $\phi$
}. $\GG_0$ is a fixed value to initialize the non-linear transform.
$\alpha$ is defined to be a scalar to adjust the lighting gain globally, and $\GG$ is defined as a three-dimensional vector for color distortion correction by applying a small permutation to $\point{\illu}$ in three color channels, under the constraint of the prior loss functions.

By applying \EQref{eq:enhance}, $\illu$ along the ray is enhanced while $\rr$ is not changed. Hence, our model can adjust the lighting and the color of the scene while preserving its geometry information. Although our model allows more complicated transformation functions to be applied, we find through experiments that \EQref{eq:enhance} works well with a good trade-off between performance and computational cost.

\begin{figure*}[t]
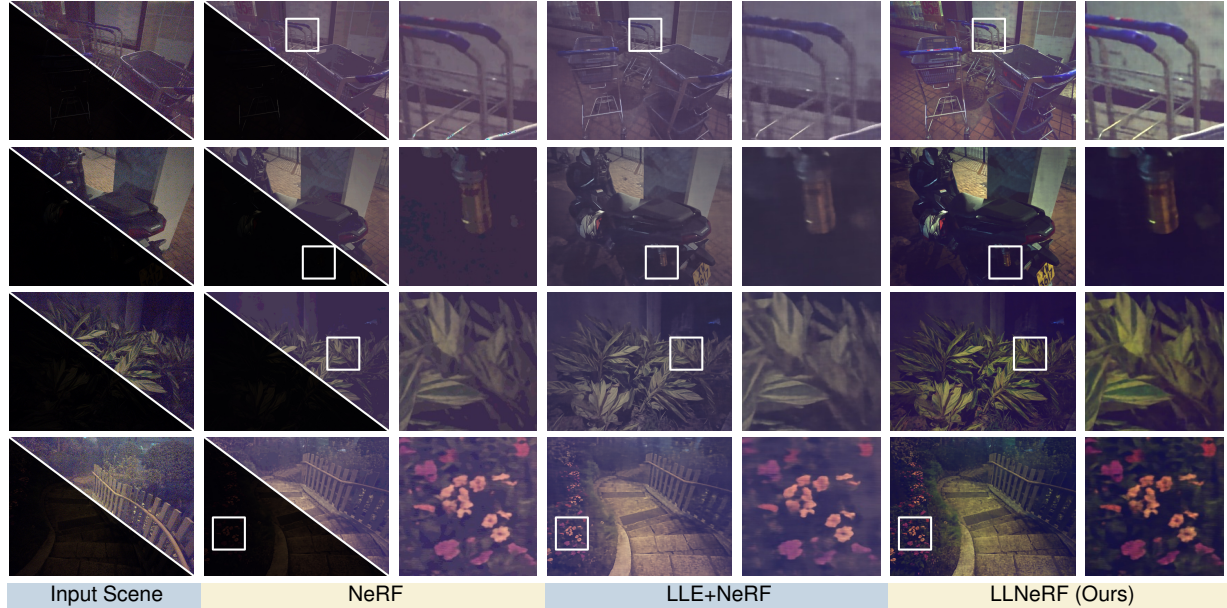

\footnotesize
\renewcommand{\tabcolsep}{1pt}
\renewcommand{\cropH}{0.105\textwidth}
\renewcommand{\addimg}[1]{
\adjustimage{height=\cropH}{images/cmp1/#1}
}
\newcommand{\cellcolorA}[1]{\cellcolor[cmyk]{0.13,0.03,0.00,0.10} #1}
\newcommand{\cellcolorB}[1]{\cellcolor[cmyk]{0.00,0.02,0.15,0.03} #1}

\centering
\begin{tabular}{c c cc cc c}

  \addimg{cmp1_1__input.png}
& \addimg{cmp1_1__nerf.png}
& \addimg{cmp1_1__nerf_patch.png}
& \addimg{cmp1_1__baseline.png}
& \addimg{cmp1_1__baseline_patch.png}
& \addimg{cmp1_1__ours_f.png}
& \addimg{cmp1_1__ours_f_patch.png}
\\
  \addimg{cmp1_2__input.png}
& \addimg{cmp1_2__nerf.png}
& \addimg{cmp1_2__nerf_patch.png}
& \addimg{cmp1_2__baseline.png}
& \addimg{cmp1_2__baseline_patch.png}
& \addimg{cmp1_2__ours_f.png}
& \addimg{cmp1_2__ours_f_patch.png}
\\
  \addimg{cmp1_3__input.png}
& \addimg{cmp1_3__nerf.png}
& \addimg{cmp1_3__nerf_patch.png}
& \addimg{cmp1_3__baseline.png}
& \addimg{cmp1_3__baseline_patch.png}
& \addimg{cmp1_3__ours_f.png}
& \addimg{cmp1_3__ours_f_patch.png}
\\
  \addimg{cmp1_4__input.png}
& \addimg{cmp1_4__nerf.png}
& \addimg{cmp1_4__nerf_patch.png}
& \addimg{cmp1_4__baseline.png}
& \addimg{cmp1_4__baseline_patch.png}
& \addimg{cmp1_4__ours_f.png}
& \addimg{cmp1_4__ours_f_patch.png}
    \\
  \multicolumn{1}{c}{\cellcolorA{\textsf{Input Scene}}}
& \multicolumn{2}{c}{\cellcolorB{\textsf{NeRF}}}
& \multicolumn{2}{c}{\cellcolorA{\textsf{LLE+NeRF}}}
& \multicolumn{2}{c}{\cellcolorB{\textsf{{\mn} (Ours)}}}
\end{tabular}
\vspace{.7em}
\caption{Visual comparison of novel view synthesis results of our model, NeRF, and the baseline model (LLE + NeRF). Note that the input scene image and the NeRF result are brightened for a better view. Our results have the best quality, with realistic color and fine details.}
\label{fig:cmp1}
\vspace{-4mm}
\end{figure*}

\begin{figure}[h]
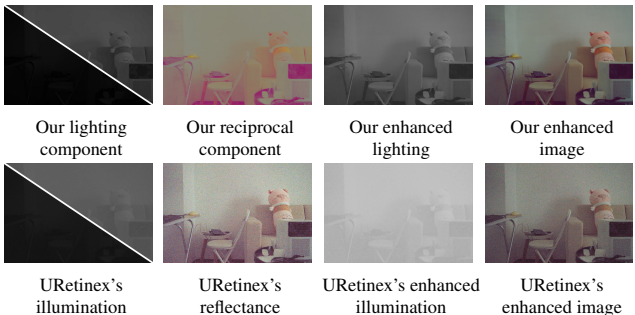

\small
\renewcommand{\tabcolsep}{1pt}
\renewcommand{\cropH}{0.075\textwidth}
\renewcommand{\addimg}[1]{
\adjustimage{height=\cropH}{images/lr-cmp/#1}
}
\centering
\begin{tabular}{cccc}
  \addimg{l_ours_tri.png}
& \addimg{r_ours.png}
& \addimg{le_ours.png} 
& \addimg{output_ours.png} 
\\

\begin{tabular}{@{}c@{}}\scriptsize{Our lighting} \\[-0.7ex] \scriptsize{component}\end{tabular} &
\begin{tabular}{@{}c@{}}\scriptsize{Our reciprocal} \\[-0.7ex] \scriptsize{component}\end{tabular} &
\begin{tabular}{@{}c@{}}\scriptsize{Our enhanced} \\[-0.7ex] \scriptsize{lighting}\end{tabular} &
\begin{tabular}{@{}c@{}}\scriptsize{Our enhanced} \\[-0.7ex] \scriptsize{image}\end{tabular}
\\

  \addimg{l_uretinex_tri.png}
& \addimg{r_uretinex.png}
& \addimg{le_uretinex.png} 
& \addimg{output.png} 
\\

\begin{tabular}{@{}c@{}}\scriptsize{URetinex's} \\[-0.7ex] \scriptsize{illumination}\end{tabular} &
\begin{tabular}{@{}c@{}}\scriptsize{URetinex's} \\[-0.7ex] \scriptsize{reflectance}\end{tabular} &
\begin{tabular}{@{}c@{}}\scriptsize{URetinex's enhanced} \\[-0.7ex] \scriptsize{illumination}\end{tabular} &
\begin{tabular}{@{}c@{}}\scriptsize{URetinex's} \\[-0.7ex] \scriptsize{enhanced image}\end{tabular}
    
\end{tabular}
\caption{\ryn{Visualization comparison on the} decomposition of our model and the 2D-based method (URetinexNet~\cite{Wu_2022_URetinex}). Dark images are brightened for a better view.
}
\label{fig:lr-cmp}
\vspace{-4mm}
\end{figure}

\subsection{Optimization Strategy}

We train our model in an end-to-end manner, as shown in \cref{fig:loss}. 
While iteratively optimizing our model across the rays of the training dataset, three kinds of supervision signals are provided: gray-world prior-based colorimetric supervision and smooth prior-based supervision are used to optimize the enhancement network, and data \ryn{supervision} is used to optimize the radiance field.

{\flushleft \bf Gray-world Prior-based Colormetric Supervision.} To correct the bias mentioned in \cref{denoising}, 
we formulate a simple but effective gray-world prior-based loss $L_c$ to constrain the learning of the enhancement network $\phi$ to produce realistic images, as:
\begin{equation}
    L_c = 
    \EE[(\pixel{\cch} - e)^2]
    + 
    \lambda_1 \EE\left[{\text{var}_{c}(\pixel{\cch}) \over \beta_1 + \text{var}_{c}(\pixel{\rr})}\right]
    + 
    \lambda_2 || \gamma ||_2
    ,
    \label{eq:color_loss}
\end{equation}
where $e, \beta_1, \lambda_1, \lambda_2$ are hyper-parameters and $\text{var}_c$ denotes the channel-wise variance. The first term \ryn{of Eq.~\ref{eq:color_loss}} is to improve the brightness of the pixels (where $e=0.55$). The second term is to correct colors based on the gray world prior, which pushes the distorted colors to the natural distribution 
by reducing the variance across three channels. To \ryn{prevent the rendered pixels from converging} to gray, we further add a dynamic weight based on the color of the weighted color basis $\rr$ along the rays to relax the constraint for highly saturated colors.
The third term is the regularization term to prevent overfitting.

{\flushleft \bf Smoothness Prior-based Supervision.} 
To preserve the color and structure of the scene in the enhanced radiance field and constrain the learning of the coefficients ($\alpha$ and $\gamma$), we expect the 
integrated coefficients
to produce locally smoothed maps.
Hence, we constrain the gradient of the weighted \ryn{sum} of these two coefficients with respect to the integrated $\pixel{\illu}$ in the image space, as:
\begin{equation}
    L_s = 
    \underbrace{
        \EE\left[
        \left({\partial \alpha_r \over \partial \pixel{\illu}}\right)^2
        \right]
    }_{L_{sa}} + 
    \underbrace{
        \EE\left[
        \left({\partial \gamma_r \over \partial \pixel{\illu}}\right)^2
        \right]
    }_{L_{sg}}.
    \label{eq:Ls}
\end{equation}
%

Since it is difficult to obtain the desired gradient information directly from \EQref{eq:Ls} due to the randomly sampled rays in training, we formulate a
discrete approximation $L_{sa}$ of \EQref{eq:Ls} as:
\begin{equation}
L_{sa} = {1 \over 2} \left[
{
    (\alpha_r - \alpha_{rh})^2 
    \over 
    (\pixel{\illu} - \illu_{rh})^2 + \epsilon_1
} + 
{ 
    (\alpha_r - \alpha_{rv})^2 
    \over 
    (\pixel{\illu} - \illu_{rv})^2 + \epsilon_1
}
\right],
\label{eq:Lsa}
\end{equation}
where $\alpha_{rh}, \alpha_{rv}, \illu_{rh}, \illu_{rv}$ are the integrated $\alpha$ and $\illu$ of neighboring rays in the horizontal and vertical directions in the image space.
To leverage the smoothness constraint, we sample rays with their neighboring rays in each optimization step, as shown in \cref{fig:loss}. $L_{sg}$ is obtained in a similar way to $L_{sa}$.

{\flushleft \bf Data Supervision.} To learn the scene geometry, we apply the data loss in \cite{rawnerf}, which is the linearization of $\EE\left[\eta(\pixel{\cgt}) - \eta(\pixel{\cc})\right]$, where $\eta(y) = \log(y + \epsilon_2)$.
Since the majority of pixels in our training images have low intensity, the tone mapping function $\eta$ is used to amplify the errors in the dark regions to facilitate the learning process.

\begin{figure}[h]
    \vspace{-3mm}
    \centering
    \makebox[\linewidth][c]{\includegraphics[width=1.0\linewidth]{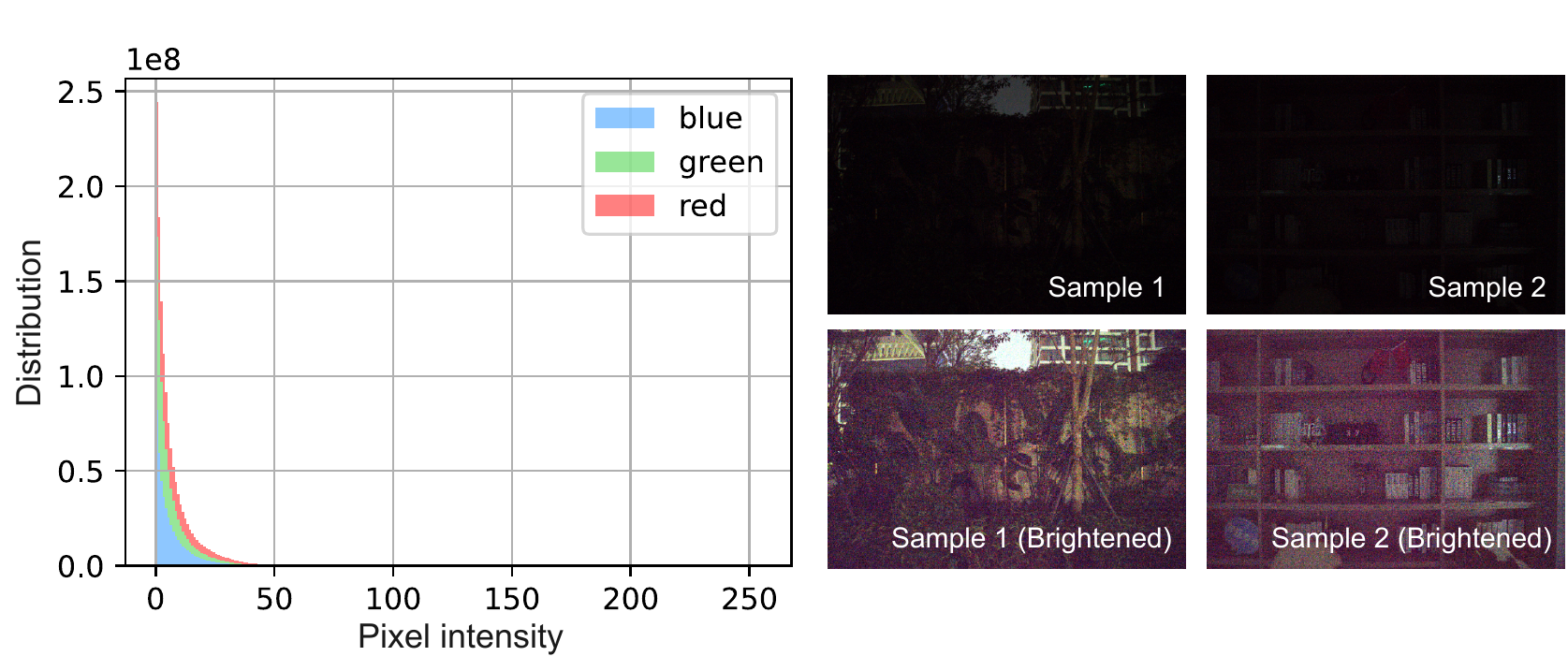}}
    \vspace{-5mm}
    \caption{
    Intensity distribution and sample images of our dataset. We collect low-light images from both indoor and outdoor scenes. These images typically have low pixel intensity, obvious color distortion, and heavy noise.
    }
    \label{fig:sample}
    \vspace{-4mm}
\end{figure}

\begin{figure*}[t]
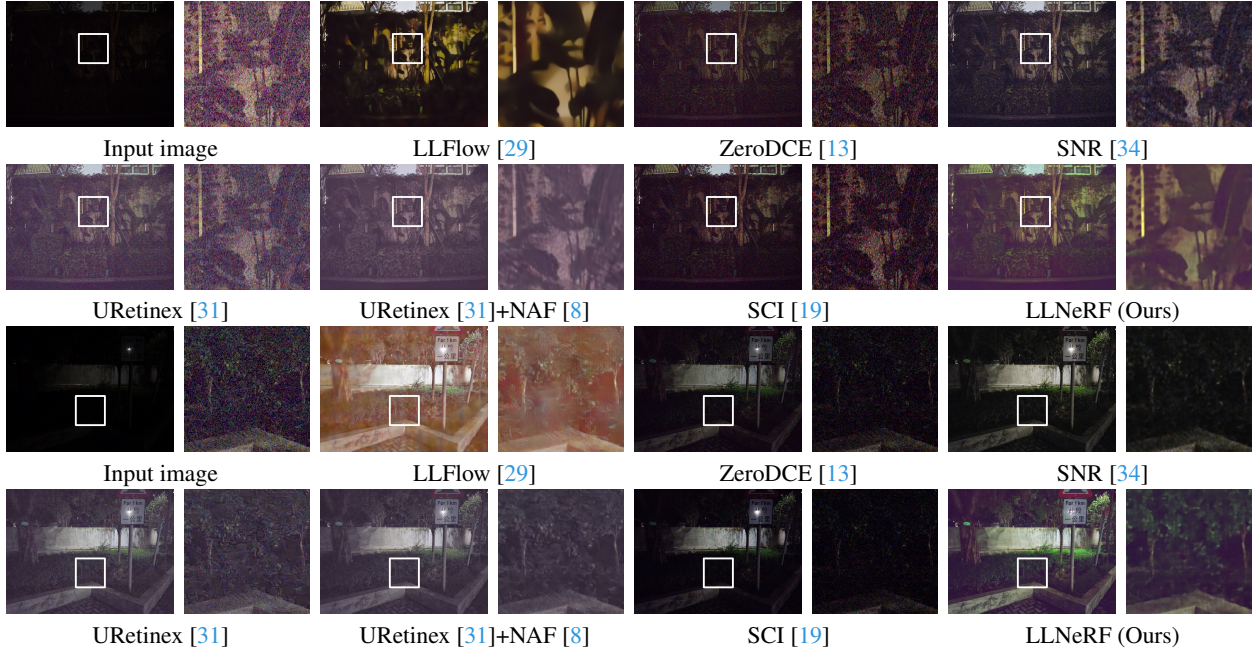

\small
\renewcommand{\tabcolsep}{1pt}
\renewcommand{\cropH}{0.095\textwidth}
\renewcommand{\addimg}[1]{
\adjustimage{height=\cropH}{images/cmp2/#1}
}
\newcommand{\lineA}{
  \multicolumn{2}{c}{Input image}
& \multicolumn{2}{c}{LLFlow~\cite{llflow}}
& \multicolumn{2}{c}{ZeroDCE~\cite{zero-dce}}
& \multicolumn{2}{c}{SNR~\cite{Xu_2022_SNRA}}
}
\newcommand{\lineB}{
  \multicolumn{2}{c}{URetinex~\cite{Wu_2022_URetinex}}
& \multicolumn{2}{c}{URetinex~\cite{Wu_2022_URetinex}+NAF~\cite{nafnet}}
& \multicolumn{2}{c}{SCI~\cite{SCI}}
& \multicolumn{2}{c}{{\mn} (Ours)}
}
\centering
\begin{tabular}{cccc cccc}
  \addimg{cmp2_1__input.png}
& \addimg{cmp2_1__input_patch.png}
& \addimg{cmp2_1__llflow_all.png}
& \addimg{cmp2_1__llflow_all_patch.png}
& \addimg{cmp2_1__dce_all.png}
& \addimg{cmp2_1__dce_all_patch.png}
& \addimg{cmp2_1__sdsd_outdoor.png}
& \addimg{cmp2_1__sdsd_outdoor_patch.png}
\\
\lineA \\
  \addimg{cmp2_1__uretinex_all.png}
& \addimg{cmp2_1__uretinex_all_patch.png}
& \addimg{cmp2_1__uretinex_naf_all.png}
& \addimg{cmp2_1__uretinex_naf_all_patch.png}
& \addimg{cmp2_1__sci.png}
& \addimg{cmp2_1__sci_patch.png}
& \addimg{cmp2_1__ours_ssvv.png}
& \addimg{cmp2_1__ours_ssvv_patch.png}
\\
\lineB \\

  \addimg{cmp2_2__input.png}
& \addimg{cmp2_2__input_patch.png}
& \addimg{cmp2_2__llflow_all.png}
& \addimg{cmp2_2__llflow_all_patch.png}
& \addimg{cmp2_2__dce_all.png}
& \addimg{cmp2_2__dce_all_patch.png}
& \addimg{cmp2_2__sdsd_outdoor.png}
& \addimg{cmp2_2__sdsd_outdoor_patch.png}
\\
\lineA \\
  \addimg{cmp2_2__uretinex_all.png}
& \addimg{cmp2_2__uretinex_all_patch.png}
& \addimg{cmp2_2__uretinex_naf_all.png}
& \addimg{cmp2_2__uretinex_naf_all_patch.png}
& \addimg{cmp2_2__sci.png}
& \addimg{cmp2_2__sci_patch.png}
& \addimg{cmp2_2__ours_f3.png}
& \addimg{cmp2_2__ours_f3_patch.png}
\\
\lineB \\
\end{tabular}
\vspace{.3em}
\caption{
The visual comparison of the results of our model and the existing low-light enhancement methods. Our results have the best quality, with realistic color and fine details.
}
\label{fig:cmp2}
\vspace{-4mm}
\end{figure*}

\section{Experiments}


\subsection{Our Dataset}
We collect a real-world dataset as a benchmark for model learning and evaluation. To obtain real low-illumination images with real noise distributions, we take photos at nighttime outdoor scenes or low-light indoor scenes containing diverse objects. Since the ISP operations are device-dependent and the noise distributions across devices are also different, we collect our data using a mobile phone camera and a DSLR camera to enrich the diversity of our dataset. We show some samples and statistics of our dataset in \cref{fig:sample}. As illustrated, the average brightness of our dataset is extremely low (most pixels' intensities are below 50 out of 255). In addition, the noise and color distortion in these images are of a very high level, making our task extremely challenging.


\subsection{Results} \label{sec:cmp}

We evaluate our model in three aspects. First, we evaluate the neural radiance field decomposition of our model by comparing to the Retinex-based state-of-the-art method URetinexNet~\cite{Wu_2022_URetinex}. Second, we evaluate the novel view synthesis performance of our model by comparing it to the baseline model (LLE + NeRF). Note that RawNeRF degrades to NeRF when RawNeRF is applied to handle sRGB images.
%
Third, we evaluate the low-light enhancement performance by comparing our model to existing state-of-the-art LLE methods.

{\flushleft \bf Visualization of $\Illu$ and $\RR$.} 
We render $\illu$ and $\rr$ via volume rendering to obtain $\Illu$ and $\RR$ for visualization, as shown in \cref{fig:loss}. \cref{fig:lr-cmp} compares our decomposition to that of URetinex~\cite{Wu_2022_URetinex}. We can see that the \hll{reflectance map} of URetinex tends to preserve all photometric information while its illumination map tends to be over-smoothed, as it is agnostic to the physical imaging process and 3D geometry information. In contrast, our model produces a more reasonable \hll{lighting-related component, and the view-independent color basis component has few shadows and lighting information.} This demonstrates the effectiveness of our decomposition \hlb{design in \cref{sec:lr}}.

{\flushleft \bf Novel View Synthesis.} 
For a fair comparison, we train our model, NeRF, and the baseline model (LLE + NeRF) using the same images and compare the novel view results, as shown in \cref{fig:cmp1}. We choose URetinexNet as the LLE model in the baseline as it tends to produce better enhancement results compared to other enhancement methods.
We can see that the results of NeRF are still low-light as there is no enhancement process inside it. Although the results of the baseline model are brightened, the image appears unrealistic as the distorted color is not corrected. In contrast, our model generates better details and natural colors.

{\flushleft \bf Low-Light Enhancement.} We further compare the results of our model with state-of-the-art low-light enhancement models. The comparison is shown in \cref{fig:cmp2}.
It shows that some methods (\ie, URetinexNet, SCI, ZeroDCE, SNR) cannot handle the noise.
While LLFlow brightens the input and removes the noise, the visual quality is still low.
We also combine the URetinexNet and a denoising model (NAFNet~\cite{nafnet} trained on SIDD~\cite{SIDD_2018_CVPR} dataset) for comparison. While this strategy can produce images with good details, the color is still distorted. In contrast, our model can enhance these images with better \hlb{cleaner} details and more natural colors.
Refer to the videos in the Supplemental for more comparisons.

\begin{figure}[!t]
\vspace{-6mm}
	\begin{center} 
		\makebox[\linewidth][c]{\includegraphics[width=1.15\linewidth]{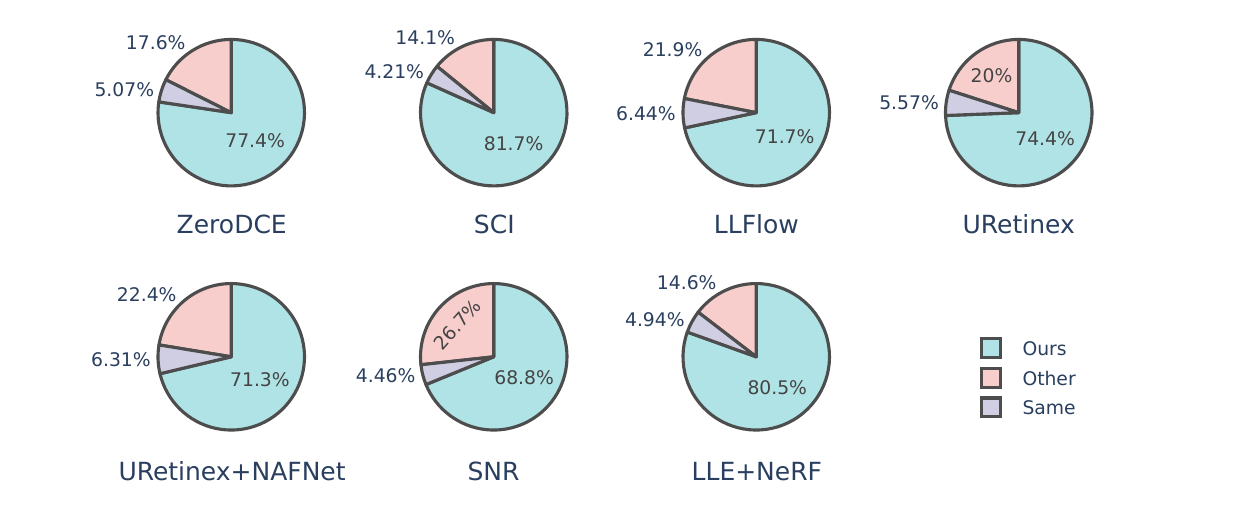}}
	\end{center}
	\vspace{-0.2in}
	\caption{
		``Ours'' is the ratio of test cases, in which the participant selected our results as better;
		``Other'' is the percentage that another method was selected to be better; and
		``Same'' is the percentage that the user has no preference.
	}
	\label{us_tbl}
\vspace{-3mm}
\end{figure}

\begin{table*}[h]

\centering
\begin{tabular}{ c|c|c|c|c|c|c } 

 \hline
 \hline
   LLE Method & LLFlow~\cite{llflow} & SNR~\cite{Xu_2022_SNRA} & SCI~\cite{SCI} & URetinex~\cite{Wu_2022_URetinex} & ZeroDCE~\cite{zero-dce} & Ours \\
 \hline
 {PSNR/SSIM}  & 16.46/0.702 & 17.04/0.575 & 12.67/0.122 & 19.18/0.289 & 13.38/0.110 & \textbf{20.50}/\textbf{0.758} \\
 \hline 
 \hline 
 NVS Method & LLFlow+NeRF
 & SNR+NeRF
 & SCI+NeRF
 & URetinex+NeRF
 & ZzeroDCE+NeRF
 & Ours \\ 
 \hline 
 {PSNR/SSIM} & 16.44/0.702 & 17.02/0.687 & 13.08/0.505 & 19.93/0.746 & 14.17/0.612 & \textbf{20.50}/\textbf{0.758} \\ 
 \hline
 \hline 
\end{tabular}
\vspace{3mm}

 \caption{The quantitative comparison results between ours and existing methods on test scenes with paired normal-light images. We compare novel view synthesis results (top row) and low-light enhancement results (bottom row). The best results are marked in {\bf bold}.} \label{psnr_tbl}
\end{table*}

{\flushleft \bf User Study. }\hll{Due to the absence of ground truths for our low-light dataset in real-world scenarios, we employ a user study to assess the visual quality of the results of different methods.} We invite 80 participants to conduct a user study to evaluate the perceptual quality of our results against those of existing approaches.
Specifically, we randomly chose 10 images from the test set for comparison with LLE methods and compare the enhanced results using an AB test.
For each test image, our produced result is ``A'' whereas the result from one of the baselines is ``B''.
Each participant would simultaneously see A and B (we avoid the bias by randomizing the left-right presentation order when displaying A and B in each AB-test task) and select one from: ``A is better'', ``B is better'', and ``I cannot decide''.
We ask the participants to make decisions based on natural brightness, rich details, distinct contrasts, vivid colors, and noise removal effects.

The comparison between ours and the baseline model, \ie, LLE + NeRF, is conducted similarly, where ``A'' and ``B" refers to the rendered videos.
For each participant, the number of tasks is 7 methods $\times$ 10 questions, $70$ in total. It takes on average around 30 minutes for each participant to complete the user study.

\cref{us_tbl} summarizes the user study results, which shows that our results are more preferred by the participants than all other competing methods.

\noindent{\bf Quantitative Evaluation.}
\hll{We additionally evaluate three scenes quantitatively with normal-light images of long exposures.
As shown in Tab.~\ref{psnr_tbl}, our method performs better than existing methods on both PSNR and SSIM. It also shows that NeRF helps enhance image structures (better SSIM), due to the implicit 3D information of its radiance field optimization process.
}

\begin{figure}[tb]
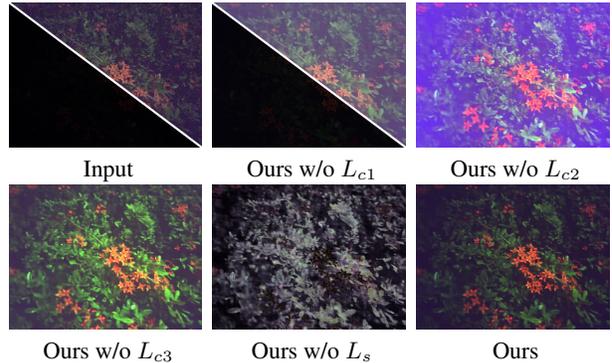

\vspace{-3mm}
\small
\renewcommand{\tabcolsep}{1pt}
\renewcommand{\cropH}{0.11\textwidth}
\renewcommand{\addimg}[1]{
\adjustimage{height=\cropH}{images/ablation/#1}
}
\centering
\begin{tabular}{ccc}
  \addimg{input_tri.png}
& \addimg{nLc1_tri.png}
& \addimg{nLc2.png} \\

Input & Ours w/o $L_{c1}$ & Ours w/o $L_{c2}$
\\

  \addimg{nLc3.png}
& \addimg{nLs.png}
& \addimg{ours.png} 
\\
Ours w/o $L_{c3}$ &  Ours w/o $L_{s}$ & Ours
    
\end{tabular}
\vspace{.3em}
\caption{
Ablation study results. $L_{c1}, L_{c2}, L_{c3}$ are three items in $L_c$ respectively. The quality of results is degraded as we remove any item. The dark images are brightened for a better view.
}
\label{fig:ablation}
\end{figure}

\noindent{\bf Ablation Study.} To investigate the effectiveness of our training strategy, we conduct the ablation study of our loss functions.
By relaxing the constraints of loss functions, we compare the visual results produced by different settings of loss functions. \cref{fig:ablation} shows that removing terms from the proposed loss function generally results in the degradation of the results produced by our model.

\begin{figure}[tb]
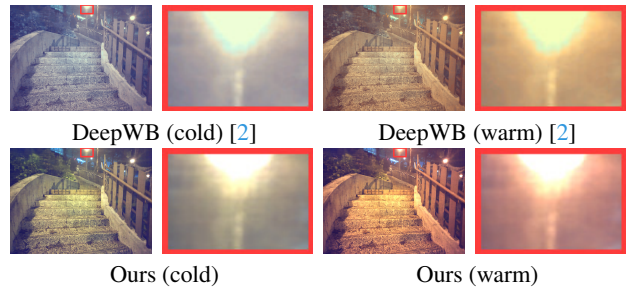

\small
\renewcommand{\tabcolsep}{1pt}
\renewcommand{\cropH}{0.08\textwidth}
\renewcommand{\addimg}[1]{
\adjustimage{height=\cropH}{images/appl/#1}
}
\centering
\begin{tabular}{cccc}
  \addimg{cmp4_1_dwb_3500.png} 
& \addimg{cmp4_1_dwb_3500_patch.png} 
& \addimg{cmp4_1_dwb_7000.png} 
& \addimg{cmp4_1_dwb_7000_patch.png} 
\\
  \multicolumn{2}{c}{DeepWB (cold)~\cite{deepWB}}
& \multicolumn{2}{c}{DeepWB (warm)~\cite{deepWB}}
\\
  \addimg{cmp4_1_cold_ours.png} 
& \addimg{cmp4_1_cold_ours_patch.png} 
& \addimg{cmp4_1_warm_ours.png} 
& \addimg{cmp4_1_warm_ours_patch.png} 
\\
  \multicolumn{2}{c}{Ours (cold)}
& \multicolumn{2}{c}{Ours (warm)}

\end{tabular}
\caption{
A possible application of our model besides the low-light enhancement. By modifying $\illu$ along the rays, our model is able to produce realistic scenes with varying color temperatures.
}
\label{fig:appl}
\vspace{-4mm}
\end{figure}

\noindent{\bf Scene Editing.} Our model allows different manipulations of the scene's illumination while producing realistic novel view images, \eg, the scene's color temperature can be edited, as shown in \cref{fig:appl}. As a comparison, the existing deep-learning-based color temperature editing method~\cite{deepWB} produces relatively unnatural editing results with artifacts in the highlight regions. 

\section{Conclusion}

In this paper, we propose a novel method to train a NeRF model from low-light sRGB images to produce novel view images of high visibility, vivid colors, and details.
Based on the observation of the imaging process, our model decomposes the neural radiance field to the lighting-related view-dependent component and view-independent color basis components in an unsupervised manner. Our model enhances the lighting without reference images under the supervision of prior-based loss functions. 
We conduct extensive experiments to analyze the properties of our method and demonstrate its effectiveness against existing methods.

{\small
\bibliographystyle{ieee_fullname}
\bibliography{egbib}
}

\end{document}